\documentclass[10pt,twocolumn,letterpaper]{article}

\usepackage{wacv}
\usepackage{times}
\usepackage{epsfig}
\usepackage{graphicx}
\usepackage{amsmath}
\usepackage{amssymb}
\usepackage{pifont}
\newcommand{\cmark}{\ding{51}}%
\newcommand{\xmark}{\ding{55}}%

\wacvfinalcopy 

\def\wacvPaperID{****} 

\ifwacvfinal\fi
\begin{document}

\title{Combinational Class Activation Maps for Weakly Supervised Object Localization}

\author{Seunghan Yang \hspace{2cm} Yoonhyung Kim \hspace{2cm} Youngeun Kim \hspace{2cm} Changick Kim\\
Korea Advanced Institute of Science Technology (KAIST), Daejeon, Korea\\
{\tt\small \{seunghan, yhkim1127, youngeunkim, changick\}@kaist.ac.kr}
}

\maketitle
\ifwacvfinal\thispagestyle{empty}\fi

\begin{abstract}
  Weakly supervised object localization has recently attracted attention since it aims to identify both class labels and locations of objects by using image-level labels. Most previous methods utilize the activation map corresponding to the highest activation source. Exploiting only one activation map of the highest probability class is often biased into limited regions or sometimes even highlights background regions. To resolve these limitations, we propose to use activation maps, named combinational class activation maps (CCAM), which are linear combinations of activation maps from the highest to the lowest probability class. By using CCAM for localization, we suppress background regions to help highlighting foreground objects more accurately. In addition, we design the network architecture to consider spatial relationships for localizing relevant object regions. Specifically, we integrate non-local modules into an existing base network at both low- and high-level layers. Our final model, named non-local combinational class activation maps (NL-CCAM), obtains superior performance compared to previous methods on representative object localization benchmarks including ILSVRC 2016 and CUB-200-2011. Furthermore, we show that the proposed method has a great capability of generalization by visualizing other datasets.
\end{abstract}

\section{Introduction}

Object localization aims to classify objects and identify their locations in a given image. Recent deep learning-based methods have demonstrated the state-of-the-art performance, especially in fully supervised settings. However, training object localization networks in a fully supervised setting requires heavy annotations, which needs a lot of time and effort to generate. Therefore, weakly supervised approaches that do not require full annotations have recently attracted attention~\cite{wei2017object, singh2017hide, zhang2018adversarial, zhang2018self}.

\begin{figure}[t]
\begin{center}
\begin{minipage}{1.0\linewidth}\vspace{0.3cm}
\centering{\epsfig{file=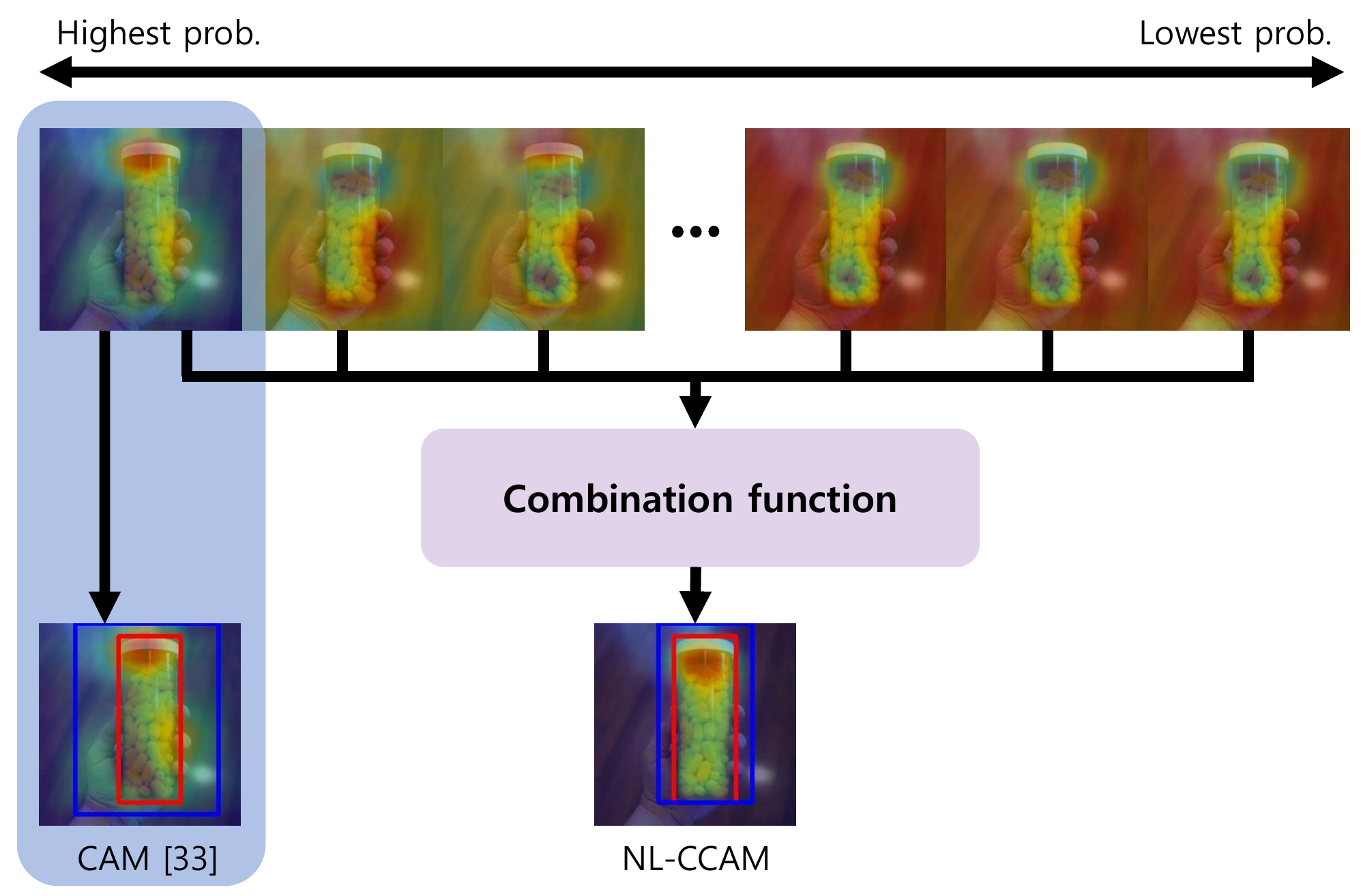, width=1.0\linewidth}}
\end{minipage}
\vspace{-2mm}
\end{center}
   \caption{Illustration of our NL-CCAM and the original CAM method. We denote the CAM method in a blue region, which uses only the activation map of the highest probability class. It unintentionally leads to highlight background regions, resulting in an inaccurate bounding box. NL-CCAM exploits all activation maps from the highest to the lowest probability class using a specific combinational function. It makes the localization map suppress background regions and highlight other parts of the object. The bounding box generated by NL-CCAM catches the target object more accurately. The predicted bounding boxes are in blue, and the ground-truth boxes are in red.}
\label{fig:long}
\label{fig:onecol}
\end{figure}

Weakly Supervised Object Localization (WSOL) is a task to identify both class labels and locations of objects in a given image by using image-level labels. It is very attractive that labeling costs for WSOL are much lower than those of fully supervised learning, which requires box-level annotations. While many CNN-based object detectors trained with full annotations surpass human performance, those trained with weak annotations still require improvements. In the case of WSOL, it is very challenging to localize objects since object locations are not given during training. Zhou \etal~\cite{zhou2014object, zhou2016learning} demonstrate that classification networks are inherently able to localize discriminative image regions without location information, and they exploit this property for WSOL. The so-called class activation maps (CAM)~\cite{zhou2016learning} inspired many researchers to adopt this concept to localize discriminative parts of an object. However, deep networks which are trained by class labels only tend to be biased on the most discriminative parts of an object. In other words, for an object in a given image, those networks tend to highlight the most discriminative portions not the whole area of the object. To overcome this limitation, recent works for WSOL aim to highlight the whole regions of an object evenly. Previous methods for this task can be categorized into the following two approaches. The first approach is to find a wide range of object parts by using spatial relationships~\cite{wei2018revisiting,zhang2018self,lee2019ficklenet}. These networks produce the state-of-the-art performance not only in WSOL but also in the weakly supervised semantic segmentation tasks. However, they only consider local relationships on high-level feature maps, resulting in coarser bounding boxes than their fully supervised counterparts, and they even tend to highlight common background regions for each class. The second approach is to erase the most discriminative parts of the object and then find new object parts~\cite{wei2017object, singh2017hide, zhang2018adversarial}. The erasing-based approach can efficiently expand discriminative parts of the object, but they often highlight regions without discriminative parts, which results in localizing common background regions. 

In this paper, to tackle the above-mentioned problems, we propose to use activation maps, named combinational class activation maps (CCAM), which are linear combinations of activation maps from the highest to the lowest probability class. To the best of our knowledge, all previous methods for WSOL exploit discriminative parts using only the activation map of the highest probability class. In contrast, we incorporate the activation maps that are formed by classes from the highest to the lowest probability. As illustrated in Fig. 1, the activation map of a higher probability class highlights some parts of the object corresponding to the class while the activation map of a lower probability class catches background regions by suppressing discriminative parts. Through empirical studies, we find that a linear combination of class activation maps have an excellent capability for suppressing background regions, and we adopt this property for WSOL. Furthermore, we design the network architecture to consider spatial relationships by using the non-local block~\cite{wang2018non}, which captures long-range dependencies via non-local operations. Specifically, unlike the previous methods that consider spatial relationships only at the high-level, we use non-local blocks at both low- and high-level layers, as shown in Fig. 2. Consideration at the low-level allows non-local use of the information such as edges and textures to capture more parts of the object when forming feature maps, and consideration at the high-level makes the network find other parts of the object associated with the most discriminative parts by using spatial and channel relations of generated feature maps. 

To summarize, we apply non-local blocks to both low- and high-level layers to find as much of the object-related regions as possible. Then, we introduce the novel algorithm to aim for the suppression of background regions, which helps to highlight foreground objects more accurately. The main contributions of this work are as follows: 

$\bullet$ We propose a novel approach to suppress background regions by using the combination of the activation maps from the highest to the lowest probability class. We also show that suppressing background regions helps to highlight the object more accurately.

$\bullet$ We propose to use non-local blocks to fit the WSOL task and localize more parts of the object considering spatial relationships of both low- and high-level feature maps.

$\bullet$ Our work achieves the state-of-the-art performance on the ILSVRC 2016 dataset with the error rate of Top-1 49.83\% and Top-5 39.31\%, and on the CUB dataset with the error rate of Top-1 47.60\% and Top-5 34.97\%.


\section{Related Work}
Recently, weakly supervised methods have received great attention in various tasks~\cite{guo2019aligned, ravanbakhsh2019training, pandey2019learning, wu2019weakly, laskar2019semantic, kolesnikov2016seed, huang2018weakly}. In this section, we review previous works by mainly focusing on WSOL. Afterwords, we also briefly introduce the non-local module~\cite{wang2018non} and review its usage for deep networks.

\subsection{Weakly Supervised Object Localization}
Weakly Supervised Object Localization (WSOL) is a challenging task to localize objects with image-level labels only.
For the first time, Zhou \etal.~\cite{zhou2016learning} show that the network learned by the classification task can be used for object localization. They obtain localization maps using the feature maps of the CAM model, which replaces the fully connected layers with a global average pooling layer since the fully connected layers of the classification network eliminate spatial information of feature maps.
Most studies on weakly supervised tasks use Zhou's method as a baseline architecture. However, since the CAM model is trained to be proficient in classification, it activates the most discriminative parts only. Recent studies have sought to find other parts of the object not just the most discriminative parts by using spatial relationships or have tried to remove the most discriminative parts and then training a classification task again to localize other parts of the object.

Researchers try to find a wide range of object parts by using spatial relationships. Wei \etal.~\cite{wei2018revisiting} propose multi dilated convolutional blocks (MDC), which can consider spatial relationships at various ratios, but not scale-invariant. Self-produced guidance (SPG) masks proposed by Zhang \etal.~\cite{zhang2018self} separate the foreground to provide the classification networks with the spatial correlation of locations, but they only consider spatial relationships locally. Lee \etal.~\cite{lee2019ficklenet} propose FickleNet, which considers spatial relationships randomly. Although this network can take into account many combinations of spatial relationships, it does not consider all possible relationships among every location. These methods of considering spatial relationships have limitations, which consider relationships locally and only exploit relationships of the high-level features. Moreover, these methods tend to highlight common background regions for each class, \eg, woods in a bird image or ocean in a ship image.

Another approach is to erase the most discriminative parts of the object and then find new object parts. Wei \etal.~\cite{wei2017object} propose the adversarial erasing (AE) network, which erases the most discriminative regions of the image to discover other parts of the object. However, this approach requires multiple classification networks. Similarly, Singh and Lee~\cite{singh2017hide} hide an image with random patches, and then seek less discriminative parts of the object but it does not consider the high-level guidance and sizes of objects. Zhang \etal.~\cite{zhang2018adversarial} propose the adversarial complementary learning (ACoL) scheme for localizing complementary object parts. They use two adversarial complementary classifiers to discover the entire objects. Although it can locate different object parts, it considers only two complementary regions that belong to the object. These erasing methods can efficiently expand discriminative parts of the object, but they often fail for images not having sufficient discriminative regions, resulting in false positives for background regions.

All of the above methods merely focus on the activation map of the highest probability class, and some of them add modules to improve the map's capability. In contrast with the previous methods, we propose to use multiple activation maps from the highest to the lowest probability class to highlight foreground objects more accurately. We show that using CCAM has a significant impact on WSOL task performance while maintaining the complexity of the network.

\subsection{Non-local Modules}
Wang \etal.~\cite{wang2018non} propose the non-local module that captures long-range dependencies directly by computing interactions between any two positions. They use non-local modules, which take both space and time, into a video classification network, and achieve the state-of-the-art performance efficiently with a slight increase in network complexity.
Recent studies have applied non-local modules to many tasks to account for long-range dependencies. Zhang \etal.~\cite{zhang2019self} propose to apply non-local modules to GAN. The attention mechanism using these modules grants more capacity for both generator and discriminator to directly model the long-range dependencies in the feature maps. As a result, the generator in~\cite{zhang2019self} can generate more realistic images than the previous methods. Zhang \etal.~\cite{zhang2019residual} first use non-local modules for image restoration to make the receptive field very large. They present very deep residual non-local attention networks using non-local modules and achieve superior performance. Moreover, many studies modify these modules for scene segmentation\cite{fu2019dual}, medical image processing~\cite{oktay2018attention}, video understanding~\cite{wu2019long,tang2018non}, graph neural network~\cite{chen2019graph}, and de-raining~\cite{li2018non}, 

It is important to find whole parts of an object in WSOL. Using long-range dependencies, a characteristic of non-local modules, the network will be able to find relevant parts of the most discriminative parts. We have applied non-local modules to the classification network to account for long-range dependencies. This is the first time to apply non-local modules in the WSOL task, and we show that our model highlights more comprehensive parts of the object than the baseline model. Furthermore, by using the proposed non-local module in conjunction with CCAM, we achieve the state-of-the-art performance on WSOL.

\section{Proposed Approach}
In this section, we first illustrate how to get the activation maps from the highest to the lowest probability class. Next, we describe our novel approach that exploits combinational class activation maps and the non-local module for WSOL.

\begin{figure*}
\begin{center}
\begin{minipage}{1.0\linewidth}
\centering{\epsfig{file=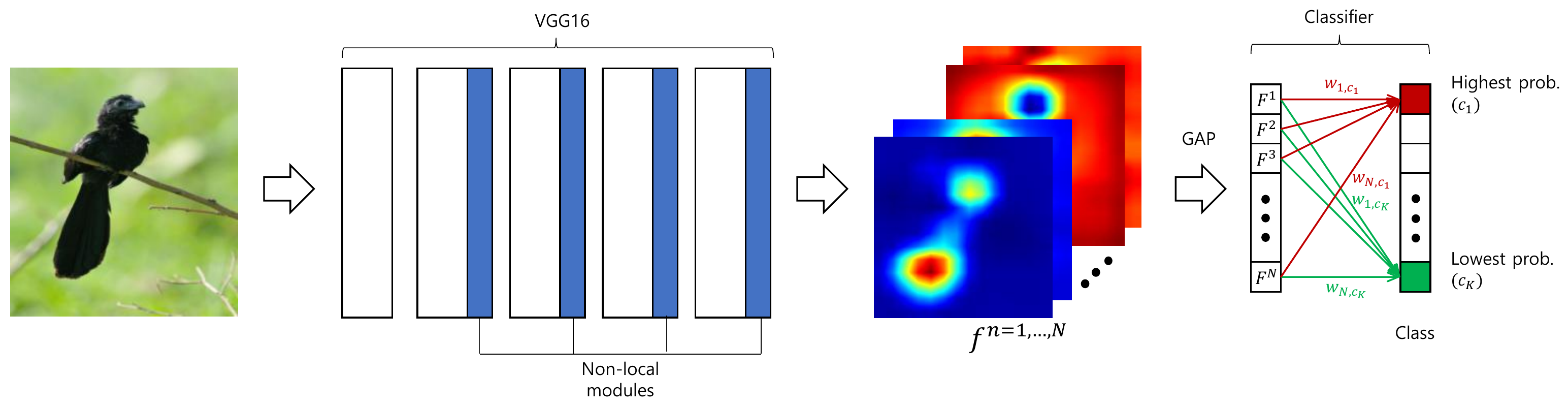,width=1.0\linewidth}}
\end{minipage}
\vspace{-2mm}
\end{center}
   \caption{The architecture of the proposed network. At inference time, we use not only the weights of the highest probability class but also those of all classes to extract activation maps from the highest to the lowest probability class.}
\label{fig:short}
\end{figure*}

\subsection{Class Activation Maps}
As mentioned earlier, the previous WSOL methods only rely on the activation map of the highest probability class. Unlike before, we observe the activation map of a higher probability class highlights some parts of the object and the activation map of a lower probability class has a considerable ability to reveal non-discriminative parts, \ie, background regions, and we utilize these properties for WSOL. To be specific, if we look into the activation map of the highest probability class, the map tends to highlight the most discriminative regions in the image. Meanwhile, if we obtain the activation map of the lowest probability class, it highlights the non-discriminative parts that are irrelevant to the object. It is because the weight parameters of the fully connected (FC) layer must be updated to include as few discriminative parts as possible in order to get the lowest probability value through the softmax layer. Therefore, the localization map created by this class highlights background regions or suppresses whole parts of the object. 

In this section, we describe how to get the activation maps from the highest to the lowest probability class, then form our final localization map. For notational convenience, we denote the activation map of the class with the $k$th highest probability as the $k$th map. In the formula, the total number of the classes is $K$, and we use ${c}_{1}, \cdots ,{c}_{K}$ to denote the class from the highest to the lowest probability in order, ${M}^{{c}_{k}}$ to denote the activation map corresponding to ${c}_{k}$. In Fig. 2, ${f}^{n}$ is the $n$th feature map before global average pooling (GAP). ${F}^{n} $ denotes the output of the GAP layer, which is the spatial average of the $n$th feature map. Here, ${F} \in {R}^{N}$, where $N$ is the channel dimension of the last feature map. At inference time, we pass the feature vector obtained by GAP to the FC layer and find the class with the highest probability through the softmax layer. The class label with the highest probability ${c}_{1}$ is given as follows:
\begin{equation}
{c}_{1} = {argmax}_{c}(\sum_{n}{w}_{n,c}{F}^{n}),
\end{equation}
where $ {w} \in {R}^{N \times K}$ indicates the weight parameter of the FC layer. 
The $1$st map, ${M}^{{c}_{1}}$ is obtained as follows:
\begin{equation}
{M}^{{c}_{1}} = \sum_{n}{w}_{n,{c}_{1}}{f}^{n}.
\end{equation}
In conjunction with ${M}^{{c}_{1}}$, the $k$th map ${M}^{{c}_{k}}$ is obtained as follows:
\begin{equation}
{M}^{{c}_{k}} = \sum_{n}{w}_{n,{c}_{k}}{f}^{n},
\end{equation}
and the final localization map is obtained by:
\begin{equation}
{M}^{ccam} = \sum_{k}g(k){M}^{{c}_{k}},
\end{equation}
where $g(k)$ is a combination function detailed in next section. As the final step, we resize $ {M}^{ccam} $ to the original input size by linear interpolation.

\subsection{Combination Functions of Activation Maps}
In Fig. 4, we visualize the activation maps from the highest to the lowest probability class in order. We observe that the $1$st map tends to highlight discriminative parts of the object, and the $K$th map highlights non-discriminative parts like background regions. Note that the $K$th map is the activation map of the lowest probability class. To use these properties effectively, we examine following two candidates for the combination function $g(k)$.

{\bf Polynomial function.}
A simple approach is to add the activation maps that catch parts of the object and subtract maps that highlight background regions. We make $g(k)$ a polynomial weight function to consider the importance of each activation map in order of probability ($g(k)$ has the largest absolute value for the $1$st map and the $K$th map). 
\begin{equation}
g(k) = \begin{cases}
\{\frac{1}{1-p}(k-p)\}^{\eta } & \text{ if } k \leq p,  \\
(-1)^{\eta+1}\{\frac{1}{p-K}(k-p)\}^{\eta } & \text{ if } k>p,
\end{cases}
\end{equation}
where $\eta$ is a degree of function and $p$ is the number of foreground activation maps. In our experiment, we set $\eta $ as $2$ to make $g(k)$ a quadratic function, and $p$ as $\frac{K+1}{2}$, where is a middle point of the number of classes.

{\bf Top-i \& bottom-j function.}
This approach only considers top-i and bottom-j class activation maps. Since not all activation maps highlight target object parts or suppress background regions, we consider only $i$ activation maps of high probability classes and $j$ activation maps of low probability classes.
\begin{equation}
g(k) = \begin{cases}
1 & \text{ if } k \leq i,  \\
-1 & \text{ if } k \geq j,  \\ 
0 & otherwise.  
\end{cases}
\end{equation}
All the previous methods only consider the activation map of the highest probability class, so we can say that they use a top-1 \& bottom-0 combination function. In our experiment, we use a top-1 \& bottom-10 combination function.

Using CCAM has several advantages for WSOL. First, there is no impact on network complexity because exploiting CCAM involves no architectural modification. The second is that introducing CCAM is free from degradation of classification performance because it is not a method of re-training a sub-network or erasing some parts when training. Existing methods~\cite{zhang2018adversarial, zhou2016learning} tend to decline classification accuracy because they see and judge remaining parts after erasing the most discriminative parts which are suitable to classify. Finally, it is possible to extract the localization map in a single forward pass at inference time while some other methods~\cite{wei2017object, lee2019ficklenet} need multiple forward passes.

\subsection{Non-local Module for WSOL}
The purpose of using spatial relationships is to look at more comprehensive areas of an object rather than to merely focus on the most discriminative parts only. Unlike previous studies~\cite{wei2018revisiting, zhang2018self, lee2019ficklenet}, which use combinations of high-level features locally, we consider spatial relationships as a non-local manner and take into account both low- and high-level feature maps simultaneously. Figure 2 shows the proposed network, and the details of composing the non-local module is motivated by~\cite{wang2018non}.

The non-local module for our work is implemented as follows. The feature maps from a certain layer $x \in {R}^{C\times H\times W}$ are first projected into three feature spaces, where $f(x), g(x) \in {R}^{C' \times H\times W }, h(x) \in {R}^{C \times H\times W}$ using $ 1\times 1$ convolution layers to embed the attention of pixels and channels. Then, we reshape $f(x)$ and $g(x)$ to ${R}^{C' \times HW}$ and $h(x)$ to ${R}^{C \times HW}$. An attention matrix is obtained as follows:
\begin{equation}
\alpha=Softmax({f({x})}^{T}g(x)), 
\end{equation}
where $\alpha \in {R}^{HW\times HW}$ indicates the weight matrix of non-local relationships, which considers the association of all pixels and channels.
In addition, we use $ 1\times 1$ convolution and batch normalization for each layer to give capacity and non-linearity, then the final attention is given as follows:
\begin{equation}
z = BN(k(h(x)\otimes\alpha)),
\end{equation}
where $BN(\cdot)$ denotes the batch normalization operation.
We add the attention layer output to the input feature map. The final output is given as follows:
\begin{equation}
y = z + x.
\end{equation} 

As illustrated in Fig. 2, we use the non-local modules described above at both low- and high-level layers. Since we consider spatial relationships at the low-level as well as the high-level, we can find more comprehensive parts of an object. Non-local blocks at the low-level help to form feature maps by combining information such as edges and textures, and those at the high-level transfer feature maps to the activation map including relevant parts of the most discriminative parts. The ablation studies in Section 4.4 shows that it is crucial to consider spatial relationships at both low- and high-level features. Finally, our network using CCAM, named NL-CCAM, can accurately highlight the object by catching more relevant parts of the object by suppressing background regions.



\section{Experiments}
In this section, we present details of experiment setups and compare our results with other methods.

\begin{figure*}
\begin{center}
\begin{minipage}{1.0\linewidth}
\centering{\epsfig{file=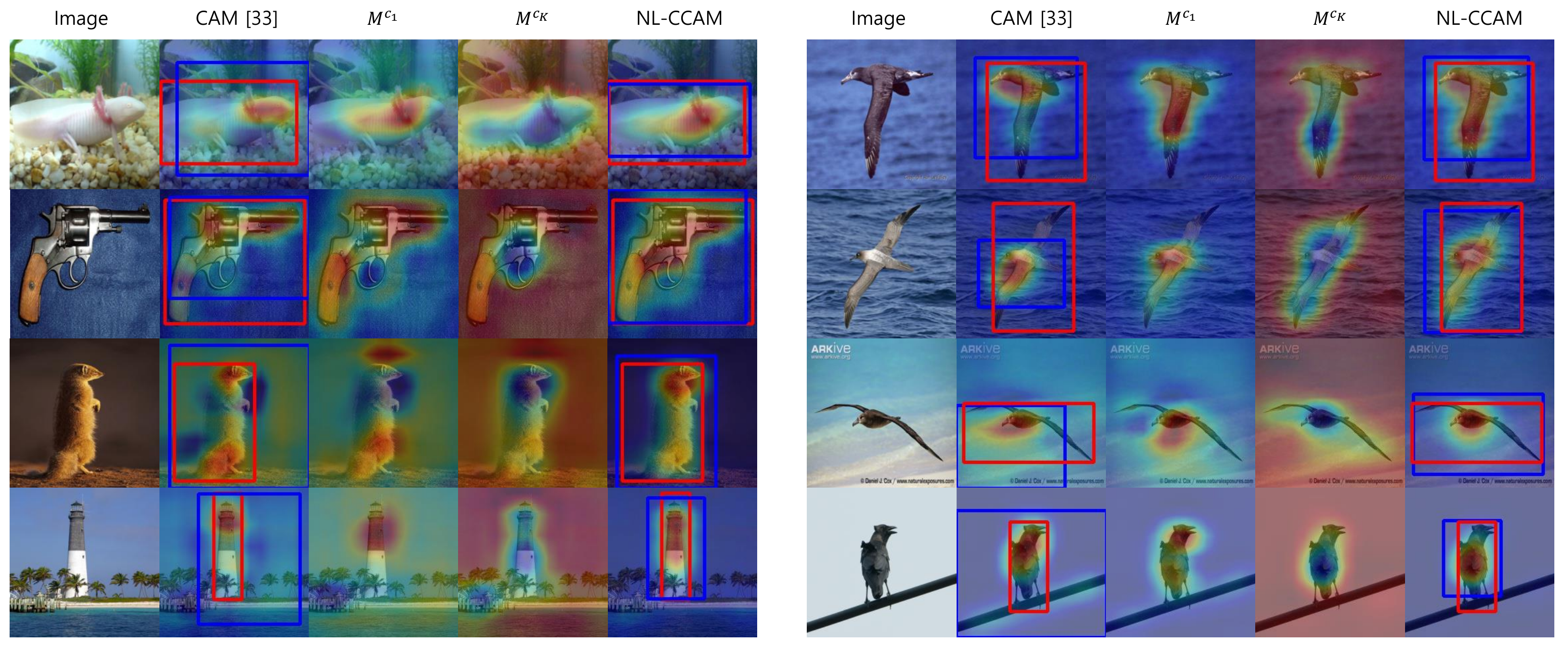,width=0.98\linewidth}}
\hspace*{0.1cm}{\hspace{0.2cm}(a) ILSVRC \hspace{6.5cm} (b) CUB-200-2011}
\vspace{-2mm}
\end{minipage}
\end{center}
   \caption{Qualitative object localization results compared with the CAM method. $M^{c_{1}}$ and $M^{c_{K}}$ stand for the $1$st map and the $K$th map , respectively, which are extracted by our network. The $1$st map catches some parts of the object while the $K$th map highlights background regions. The predicted bounding boxes are in blue, and the ground-truth boxes are in red. In $1$st and $2$nd rows, NL-CCAM ($5$th column) catches more parts of the object than the CAM model ($2$nd column), which highlights small part of the object. In $3$rd and $4$th rows, the CAM method cannot suppress background regions ($2$nd column) and cause to generate a bounding box inaccurately. NL-CCAM, on the other hand, suppresses background regions and hits a bounding box correctly ($5$th column). Best viewed in color.}
\label{fig:short}
\vspace{-1mm}
\end{figure*}

\subsection{Experiment Setup}
{\bf Datasets and evaluation metrics.} We compare the results of our NL-CCAM with baselines and the state-of-the-art approaches on two object localization benchmarks, \ie, ILSVRC 2016~\cite{russakovsky2015imagenet} and CUB-200-2011~\cite{wah2011caltech}. ILSVRC 2016 contains 1.2 million images of 1,000 categories for training. We report the accuracy on the validation set which has 50,000 images. CUB-200-2011 is a fine-grained bird dataset of 200 categories, which contains 5,994 images for training and 5,794 for testing. We use three evaluation metrics to measure performance, which are suggested by~\cite{russakovsky2015imagenet}. The first metric is {\it Classification accuracy} which judges the answer as correct when the estimated class is equal to the ground truth class. The second metric is {\it Localization accuracy} which counts a test image as a correct one when both its class label and bounding box are correctly identified. Here, a correct bounding box indicates that a predicted bounding box has more than 0.5 overlap with the ground truth. The third metric is {\it GT-known localization accuracy} which examines the bounding box correctness only under the condition that the ground truth label is given. In the supplementary materials, we also visualize the localization maps on STL-10~\cite{coates2011analysis}, Stanford-Dogs~\cite{khosla2011novel}, and Stanford-Cars~\cite{KrauseStarkDengFei-Fei_3DRR2013} to prove our approach is applicable to any dataset.

{\bf Implementation details.} We adopted the VGGnet-GAP ~\cite{zhou2016learning} as the backbone network, and composed our VGGnet-CCAM to have the same architecture with VGGnet-GAP for fair comparison. In our NL-CCAM, we inserted non-local blocks before every bottleneck layer excluding the first bottleneck layer. The backbone network was pre-trained on ILSVRC, and the newly added blocks are randomly initialized except for the batch normalization layers in the non-local modules, which are initialized as zero. We fine-tuned our network with learning rate 0.0001, batch size 32, and 30 epochs. For fair comparison, we trained and tested our network in the same way as the baseline methods~\cite{zhou2016learning, singh2017hide, zhang2018adversarial, zhang2018self}. Specifically, for training, input images were reshaped to $256 \times 256$, followed by random cropping $224 \times 224$. At test time, we resized images to $ 224 \times 224 $ directly in order to find the whole objects. To get the localization map, we selected a top-1 \& bottom-10 function and a quadratic function for ILSVRC and CUB-200-2011, respectively. We study how to choose a combination function for each dataset in section 4.3. Finally, we use the simple thresholding technique proposed by~\cite{zhou2016learning} to generate a bounding box from the localization map.

\begin{table}
\begin{center}
\begin{tabular}{|l|c|}
\hline
Methods & Top-1 err. \\
\hline\hline
VGGnet-GAP~\cite{zhou2016learning} & 33.4 \\
VGGnet & 31.2 \\
VGGnet-ACoL~\cite{zhang2018adversarial} & 32.5 \\
\hline
VGGnet-CCAM (ours) & 33.4\\
NL-CCAM (ours) & \bf 27.7\\
\hline
\end{tabular}
\end{center}
\caption{Classification errors on the ILSVRC validation set.}
\vspace{-1mm}
\end{table}

\subsection{Comparison with the State-of-the-Arts}
We report not only NL-CCAM but also VGGnet-CCAM to observe the effect of CCAM without non-local modules.

{\bf Classification.} Table 1 shows the Top-1 classification errors on the ILSVRC validation set. Since non-local modules catch more information between locations regardless of distance, our NL-CCAM achieves better classification performance than previous methods. As illustrated in Table 2, our NL-CCAM achieves the Top-1 error of 26.6\% on the CUB-200-2011 dataset without using the bounding box annotation. While some networks, \eg, VGGnet-GAP~\cite{zhou2016learning} and VGGnet-ACoL~\cite{zhang2018adversarial}, cause classification degradation by modifying the network architecture for localization, whereas our network tends to improve classification performance by adding non-local blocks.

{\bf Localization.} Localization errors on the ILSVRC validation data is shown in Table 3. We observe that VGGnet-CCAM outperforms VGGnet-GAP by 5.42\% in the Top-1 error and also shows 2.39\% better performance than VGGnet-ACoL, which uses two parallel-classifiers for discovering complementary object regions. This result shows that only using CCAM can catch the object more accurately. Furthermore, our NL-CCAM achieves 49.83\% of the Top-1 localization error, which is the new state-of-the-art result. We illustrate the localization errors on the CUB-200-2011 dataset in Table 4. Our methods are significantly better than the state-of-the-art methods. VGGnet-CCAM already outperforms the other previous methods only with the use of CCAM (\ie, without using the non-local module). Our NL-CCAM performs 5.76\% and 7.31\% points better than SPG on Top-1 and Top-5 errors. In conjunction with background suppression, considering non-local relationships at both low- and high-level feature maps leads to a powerful performance in WSOL.

\begin{table}
\begin{center}
\begin{tabular}{|l|c|c|}
\hline
Methods & Anno. & Top-1 err. \\
\hline\hline
GoogLeNet-GAP on full image~\cite{zhou2016learning} & n/a & 37.0\\
GoogLeNet-GAP on crop~\cite{zhou2016learning} & n/a & 32.2\\
GoogLeNet-GAP on BBox~\cite{zhou2016learning} & BBox & 29.5\\
VGGnet-ACoL~\cite{zhang2018adversarial} & n/a & 28.1 \\
\hline
VGGnet-CCAM (ours) & n/a & 26.8\\
NL-CCAM (ours) & n/a & \bf 26.6\\
\hline
\end{tabular}
\end{center}
\caption{Classification errors on the CUB-200-2011 test set.}
\vspace{-1mm}
\end{table}

Furthermore, we compare the GT-known localization errors to eliminate the influence caused by classification results. Table 5 shows that NL-CCAM achieves 34.77\% in the Top-1 error on the ILSVRC validation set. It means that the proposed method generates the localization map more accurately regardless of classification results. 
We also compare the GT-known localization errors on the CUB-200-2011 dataset in the supplementary materials.

\begin{table}
\begin{center}
\begin{tabular}{|l|c|c|}
\hline
Methods & Top-1 err. & Top-5 err.\\
\hline\hline
AlexNet-GAP~\cite{zhou2016learning} & 67.19 & 52.16 \\
Backprop on GoogLeNet~\cite{simonyan2013deep} & 61.31 & 50.55 \\
GoogLeNet-GAP~\cite{zhou2016learning} & 56.40 & 43.00 \\
GoogLeNet-HaS-32~\cite{singh2017hide} & 54.79 & - \\
GoogLeNet-ACoL~\cite{zhang2018adversarial} & 53.28 & 42.58 \\
Backprop on VGGnet~\cite{simonyan2013deep} & 61.12 & 51.46 \\
VGGnet-GAP~\cite{zhou2016learning} & 57.20 & 45.14 \\
VGGnet-ACoL~\cite{zhang2018adversarial} & 54.17 & 40.57 \\
SPG-plain~\cite{zhang2018self} & 53.71 & 41.81 \\
SPG~\cite{zhang2018self} & {51.40} & {40.00} \\
\hline
VGGnet-CCAM (ours) & 51.78 & 40.64\\
NL-CCAM (ours) & \bf {49.83} & \bf {39.31}\\
\hline
\end{tabular}
\end{center}
\caption{Localization errors on the ILSVRC validation set.}
\end{table}

\begin{table}
\begin{center}
\begin{tabular}{|l|c|c|}
\hline
Methods & Top-1 err. & Top-5 err.\\
\hline\hline
GoogLeNet-GAP~\cite{zhou2016learning} & 59.00 & - \\
VGGnet-ACoL~\cite{zhang2018adversarial} & 54.08 & 43.49 \\
SPG-plain~\cite{zhang2018self} & 56.33 & 46.47 \\
SPG~\cite{zhang2018self} & 53.36 & 42.28 \\
\hline
VGGnet-CCAM (ours) & {49.93} & {36.25}\\
NL-CCAM (ours) & \bf {47.60} & \bf {34.97}\\
\hline
\end{tabular}
\end{center}
\caption{Localization errors on the CUB-200-2011 test set.}
\vspace{-1mm}
\end{table}

{\bf Visualization.} Figure 3 shows activation maps and bounding boxes of the CAM method and proposed method on ILSVRC and CUB-200-2011. We visualize a map by CAM and three maps extracted by our network. We first show our proposed network catches more comprehensive parts of the object than the CAM model (compared with CAM and $M^{c_{1}}$ in Fig. 3). For example, in the results of CAM, only the most discriminative parts are highlighted, \eg, the muzzle of the gun or the face of the bird, whereas our model can find more relevant parts, \eg, the handle of a gun or the wings of the bird. However, both CAM and the $1$st map, which use the activation map of the highest probability class, tend to highlight common background regions, \eg, a wood or sky in a bird image. NL-CCAM suppresses background regions thoroughly and highlights more parts of the object. Interestingly, suppressing background regions catches the unhighlighted portions of the $1$st map, which leads to locate the object accurately.

\begin{table}
\begin{center}
\begin{tabular}{|l|c|}
\hline
Methods & GT-known loc. err. \\
\hline\hline
AlexNet-GAP~\cite{zhou2016learning} & 45.01 \\
AlexNet-HaS~\cite{singh2017hide} & 41.25 \\
AlexNet-GAP-ensemble~\cite{zhou2016learning} & 42.98 \\
AlexNet-HaS-ensemble~\cite{singh2017hide} & 39.67 \\
GoogLeNet-GAP~\cite{zhou2016learning} & 41.34 \\
GoogLeNet-HaS-32~\cite{singh2017hide} & 39.71 \\
Deconv~\cite{zeiler2014visualizing} & 41.60 \\
Feedback~\cite{cao2015look} & 38.80 \\
MWP~\cite{zhang2018top} & 38.70 \\
ACoL~\cite{zhang2018adversarial} & 37.04 \\
SPG-plain~\cite{zhang2018self} & 37.32 \\
SPG~\cite{zhang2018self} & {35.31} \\
\hline
VGGnet-CCAM (ours) & 36.42 \\
NL-CCAM (ours) & \bf {34.77}\\
\hline
\end{tabular}
\end{center}
\caption{GT-known localization errors on the ILSVRC validation set.}
\end{table}


\begin{table}
\begin{center}
\begin{tabular}{|c|c|c|c|}
\hline
A combination function & ILSVRC & CUB-200-2011 \\
\hline\hline
top-1 \& bottom-0 & 54.17 & 50.55 \\
\hline
top-0 \& bottom-1 & 52.30 & 52.71 \\
\hline
top-1 \& bottom-1 & 50.77 & 49.19 \\
\hline
top-1 \& bottom-10 & \bf{49.83} & 48.07\\
\hline
top-1 \& bottom-20 & 49.90 & 48.41\\
\hline
Constant ($\eta=0$) & 52.91 & 47.77 \\
\hline
Linear ($\eta=1$) & 52.71 & 47.64 \\
\hline
Quadratic ($\eta=2$) & 52.57 & \bf{47.60} \\
\hline
Cubic ($\eta=3$) & 52.21 & 47.64 \\

\hline
\end{tabular}
\end{center}
\caption{The effect of a combination function.}
\vspace{-1mm}
\end{table}

\begin{figure*}
\begin{center}
\begin{minipage}{1.0\linewidth}
\centering{\epsfig{file=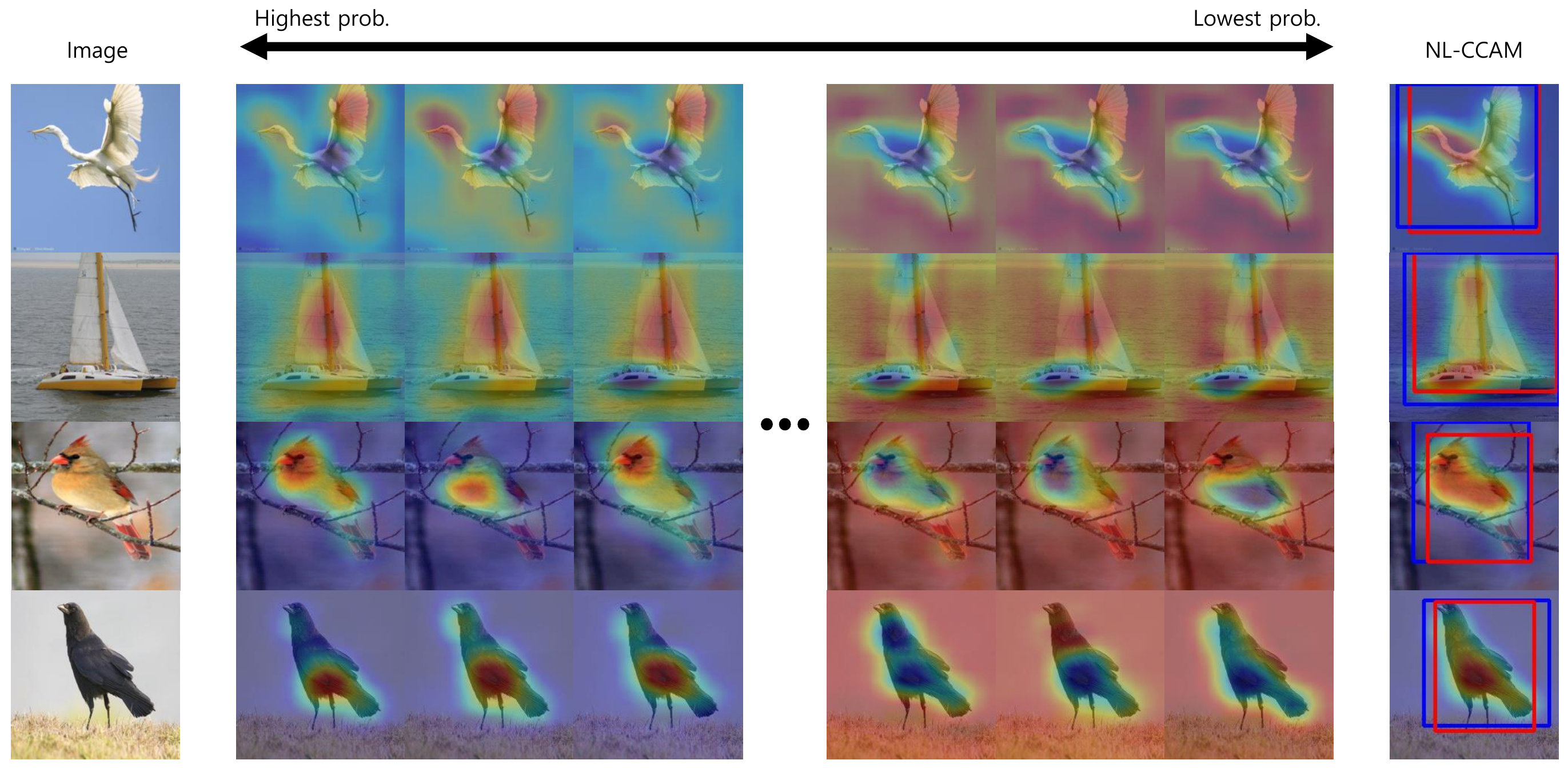,width=0.98\linewidth}}
\end{minipage}
\vspace{-3mm}
\end{center}
   \caption{The activation maps of the proposed method on ILSVRC and CUB-200-2011. The first two rows are the activation maps on ILSVRC, and the last two rows are the activation maps on CUB-200-2011. For each image, we illustrate the first three maps and the last three maps (middle column) and NL-CCAM (last column). The activation maps of higher probability classes tend to catch the parts of the object, and the maps of lower probability classes tend to highlight the background regions. The predicted bounding boxes are in blue, and the ground-truth boxes are in red.}
   \vspace{-1.5mm}
\label{fig:short}
\end{figure*}

\subsection{The Choice of a Combination Function}
In Fig. 4, we visualize the first three activation maps and the last three activation maps. Based on our observation of these activation maps, we select a specific function for each dataset. In table 6, using a top-1 \& bottom-0 combination function, which is used in all previous studies, achieves lower performance than using other functions. First, we show that using a top-0 \& bottom-1 combination function, which makes the localization map using only the inverted $K$th map, performs 52.30\% on the ILSVRC dataset. As a result of this experiment, we demonstrate that the $K$th map has a capability of suppressing background regions. To take advantage of this property, we experimented with the simplest case, top-1 \& bottom-1, which uses only the $1$st map and the $K$th map. It already outperforms the previous state-of-the-art method by 0.63\% and 4.17\% on ILSVRC and CUB-200-2011. On the ILSVRC dataset, using a top-1 \& bottom-10 function achieves the best performance as 49.83\%, but using polynomial combination functions that exploit all activation maps achieves relatively low performance. It is because the activation map of a high probability class highlights object parts corresponding to that class not the target class. For example, the $2$nd map in Fig. 1 highlights parts of a hand, which do not belong to the target object. Therefore using multiple activation maps of high probability classes may cause degradation of localization accuracy by highlighting other objects. On the contrary, using polynomial combination functions leads to a great performance on the CUB-200-2011 dataset. This is because this dataset consists of 200 classes of birds, which are relevant to each other. In this case, as illustrated in Fig. 4, the activation maps of higher probability classes highlight some parts of the bird and the combination of all activation maps helps to localize the object entirely.

\begin{table}
\begin{center}
\begin{tabular}{|c|c|c|c|}
\hline
Non-local & Non-local & CCAM & Top-1 err. \\
at the low-level & at the high-level & & \\
\hline\hline
\xmark & \xmark & \xmark & 57.84 \\
\hline
\cmark & \xmark & \xmark & 55.13 \\
\hline
\xmark & \cmark & \xmark & 56.94 \\
\hline
\cmark & \cmark & \xmark & 50.55 \\
\hline
\xmark & \xmark & \cmark & 49.93 \\
\hline
\cmark & \xmark & \cmark & 48.41 \\
\hline
\xmark & \cmark & \cmark & 49.19 \\
\hline
\cmark & \cmark & \cmark & \bf 47.60 \\

\hline
\end{tabular}
\end{center}
\caption{Ablation studies on the CUB-200-2011 test set.}
\vspace{-1mm}
\end{table}

\subsection{Ablation Studies}
To better understand the effectiveness of each proposed module, we conducted several ablation studies. 

{\bf Non-local modules.} The results of our ablation studies for the Top-1 error on CUB-200-2011 are illustrated in Table 7. We observe that using non-local modules at both low- and high-level layers leads to a big boost as almost 5\% compared to using them only at the low-level or at the high-level. This result shows that the non-local relationships at the low-level help to localize more parts of the object when considering the non-local relationships at the high-level to form feature maps.

{\bf Single (CAM) vs. multiple (CCAM) activation maps.} The use of CCAM shows substantial performance improvements without additional networks. In particular, the performance is increased by 7.91\% compared to the baseline by only using CCAM. Furthermore, regardless of where non-local modules are used, exploiting CCAM performs better than original models. In this result, the localization map using CCAM can suppress background regions well, and we show that background suppression is as essential as finding the whole parts of the object.


\section{Conclusion}
In this paper, we have proposed NL-CCAM for localizing object regions in WSOL. We first adapt non-local modules to WSOL and improve classification and localization performance on ILSVRC 2016 and CUB-200-2011. Moreover, we observe the activation maps from the highest to the lowest probability class and the combination of these maps having a great ability to reveal non-discriminative parts. We utilize this property to suppress background regions, resulting in precise localization of the object. Extensive experiments show the proposed method can localize more object regions on multiple datasets and outperform the previous state-of-the-art methods.

{\small
\bibliographystyle{ieee}
\bibliography{egbib}
}

\end{document}